\documentclass[10pt,letterpaper]{article}

\usepackage{cogsci}

\cogscifinalcopy 

\usepackage{multirow}
\usepackage{color}
\usepackage{verbatim}
\usepackage{makecell}
\usepackage{pslatex}
\usepackage{apacite}
\usepackage{float} 
\usepackage{graphicx}

\title{CAUS: A Dataset for Question Generation based on Human Cognition\\ Leveraging Large Language Models}
 
\author{{\large \bf Minjung Shin (mjshin77@snu.ac.kr)} \\
  Interdisciplinary Program in Cognitive Science, Seoul National University \\
  Gwanak-ro, Gwanak-gu, Seoul, 08826, Republic of Korea\\
  \AND {\large \bf Donghyun Kim (kimdonghyun0704@gmail.com), Jeh-Kwang Ryu (ryujk@dgu.ac.kr)} \\
  Department of Physical Education, Dongguk University \\
  Pildong-ro, Jung-gu, Seoul, 08826, Republic of Korea}

\begin{document}

\maketitle

\begin{abstract}

We introduce the \textbf{C}urious \textbf{A}bout \textbf{U}ncertain \textbf{S}cene (\textbf{CAUS}) dataset, designed to enable Large Language Models, specifically GPT-4, to emulate human cognitive processes for resolving uncertainties. Leveraging this dataset, we investigate the potential of LLMs to engage in questioning effectively. Our approach involves providing scene descriptions embedded with uncertainties to stimulate the generation of reasoning and queries. The queries are then classified according to multi-dimensional criteria. All procedures are facilitated by a collaborative system involving both LLMs and human researchers. Our results demonstrate that GPT-4 can effectively generate pertinent questions and grasp their nuances, particularly when given appropriate context and instructions. The study suggests that incorporating human-like questioning into AI models improves their ability to manage uncertainties, paving the way for future advancements in Artificial Intelligence (AI). 

\end{abstract}

\section {Introduction}

The significance of questioning, rather than just answering, lies in the entity's latent capacity to seek information, regardless of whether the entity is a human or a machine.
While questioning as a learning strategy comes naturally to humans, its implementation in machines, i.e., question generation (QG), is relatively recent \cite{duan2017question, chen2018learningq, zhou2019question, gong2022khanq}.
Although QG models have yielded positive results in active learning \cite{misra2018learning, Krishna2022socialAI} and user engagement\cite{huang2017doesn}, this topic remains on the fringes rather than a mainstream of artificial intelligence (AI).

In recent years, AI models, especially Large Language Models (LLMs) have garnered significant attention for their proficient language generation. 
These models can even engage in zero-shot learning, generating text for scenarios not covered in their training data \cite{brown2020language, wei2022emergent}.
However, closer examination reveals significant shortcomings in LLMs, particularly in planning tasks when subjected to systematic evaluations. While competent with surface-level structures, they struggle in deeper planning \cite{momennejad2023evaluating}. 
Furthermore, while they outperform humans in generating text and images, their understanding performances are not as robust \cite{west2023generative}. 
These issues become evident when user instructions are ambiguous, leading to inconsistent or erratic responses, such as hallucinations \cite{gallegos2023bias} or vulnerabilities \cite{wang2023can}.
The main issue with LLMs is that they are often deployed to address inference tasks based on probabilistic contexts without engaging in follow-up questions, even in uncertain situations \cite{toles2023good}.

Then, are LLMs inherently incompetent at asking questions? 
Can't we improve AI models by implementing human questioning strategies to deal with uncertainty? 
With these questions in mind, we propose a text dataset named \textbf{CAUS} (\textbf{C}urious \textbf{A}bout \textbf{U}ncertain \textbf{S}cene) that emulates human cognitive processes for resolving uncertainty, with reasoning and asking questions. The main contributions of our hypothetical approach are as follows:

- Providing a diverse range of questions proper to specific scenarios.\\
\indent - Classifying questions based on multi-dimensional criteria, considering their attributes, scope, and format.\\
\indent - Establishing a collaborative system between LLMs and human researchers to enhance efficiency and relevance.

In our study, we focus on \textit{sincere information-seeking questions}, which aim to resolve uncertainty in a given situation \cite{flammer1981towards, graesser2003does}.
By excluding social interaction elements like persuasive or request questions, we maintain a clear information-seeking intent.
Our approach assumes a scenario where an agent asks an oracle to obtain specific information, thus resolving uncertainty.

We employ \textit{Scene Description} texts as the starting point for question generations.
For each scene, we produce \textit{Reasoning}, which captures how humans clarify uncertain entities, and \textit{Questioning}, which generates relevant questions to resolve uncertainties.
Concurrently, we classify the \textit{Type of Generated Question} as a basis for systematic question evaluation. 
The constructed dataset, along with the code and prompts, are distributed for further use \footnote{https://github.com/lbaa2022/CAUS\_v1}.

\section{Related Work}

\subsection{Research on Questioning}
Research on asking questions is relatively scarce compared to answering them. This scarcity is due to two key issues: 1) difficulties with defining the scope of various questions (e.g., rhetorical questions, questions given in plain text) and 2) various contexts in which they occur (e.g., requests, social coordination, expressing complaints) \cite{graesser1985psychology}.
Although research on questioning is limited, there is consensus that the cognitive process underlying questioning is \textit{cognitive disequilibrium} and \textit{the questioner's desire to resolve it} \cite{flammer1981towards, otero2001preg, graesser2003does, loewenstein1994psychology}. 
The cognitive disequilibrium stems from knowledge gaps, anomalies, contradictions, discrepancies, unexpected outcomes, or goal-blocking obstacles \cite{graesser2013mechanisms}.
The questioner's desire refers to humans' robust motivation for information exploration, i.e., curiosity  \cite{golman2018desire, Vazard2022noetic}.

Empirical studies, especially in education, underscore the value of effective questioning in enhancing learning, advocating for the promotion of students' questioning skills  \cite{graesser1993anomalous, rosenshine1996teaching, macagno2023questions}. 
Outside of educational contexts, studies are primarily conducted in gaming, while quantifying uncertainty in open domains presents significant challenges. 
These studies indicate that adept questioners often demonstrate strong strategic thinking, but effective questioning is not an easy feat, even for humans \cite{rothe2018people}.
However, even if people usually ask inefficient questions, asking behavior itself plays an essential role in learning \cite{Cervera2020SysNeuro} and desirable development \cite{chouinard2007children}.
To sum up, asking questions is crucial but challenging when formulating relevant ones.

\subsection{Good Questioning} 

Most discussions on \textit{what makes a good question} center around \textit{learning}, reflecting the inquisitive nature of asking questions.
In educational contexts, good questions encourage deep reasoning and active exploration.
Inquiries, such as ``why," ``how," ``what if," and ``what if not," are valuable because they delve into causal, goal-oriented, and logical reasoning \cite{graesser2003does, Graesser2009whatGQ}. Such questions are linked to understanding, problem-solving, reasoning, creativity, and other cognitive processes. They encourage learners to engage more profoundly with the material, enhancing learning and literacy \cite{macagno2023questions, otero2001preg}.

Finding a solid and nominal criterion for good questions is challenging outside of pedagogy. 
Indeed, in the everyday life of humans, defining the exact and acute question is not always significant. However, defining what constitutes good or bad questions in the context of QG is essential. 
Thus, individual research efforts often define their own criteria to evaluate questions, reflecting the complex and diverse requirements of questioning tasks in different contexts.

In Battleship, a strategy guessing game, good questions target valuable information about the hidden configuration of the game board. The question should be specific to the context and aimed at resolving uncertainties about ship size and position \cite{Rothe2017NIPS}.
A proposal of the Questioning Turing Test emphasizes that questioning provides a more nuanced measure of AI than passive responding. The study suggests three evaluation criteria: \textit{human-likeness}, \textit{correctness} (i.e., whether the entity fulfills the inquiry), and \textit{strategicness} (i.e., accessing goal by fewer questions) \cite{damassino2020QTT}.

Among various criteria, \textit{question diversity} is underscored as a beneficial index of good questioning in improving QA results and user engagement. 
Diverse questions can cover broader topics, content, and difficulty levels, which is crucial for comprehensive understanding and assessment in learning contexts \cite{sultan2020diverseQG}. 
Studies have shown that when questions have varied types, syntax, and content, they require diverse answers, enhancing learning and evaluation processes \cite{yoon2023storybook}. 
Another study revealed the number of unique questions and novelty are positively related to the performance of a visual QG system regarding engagement and effectiveness \cite{Jain2017CVPR}.
However, evaluation criteria for existing research tend to be either excessively mechanical (e.g., automatic scoring based on similarity) or exceedingly subjective (e.g., ranked by human annotators).

\section {Dataset Design}
\begin{figure}[h]
\begin{center}
\includegraphics[width=\linewidth]{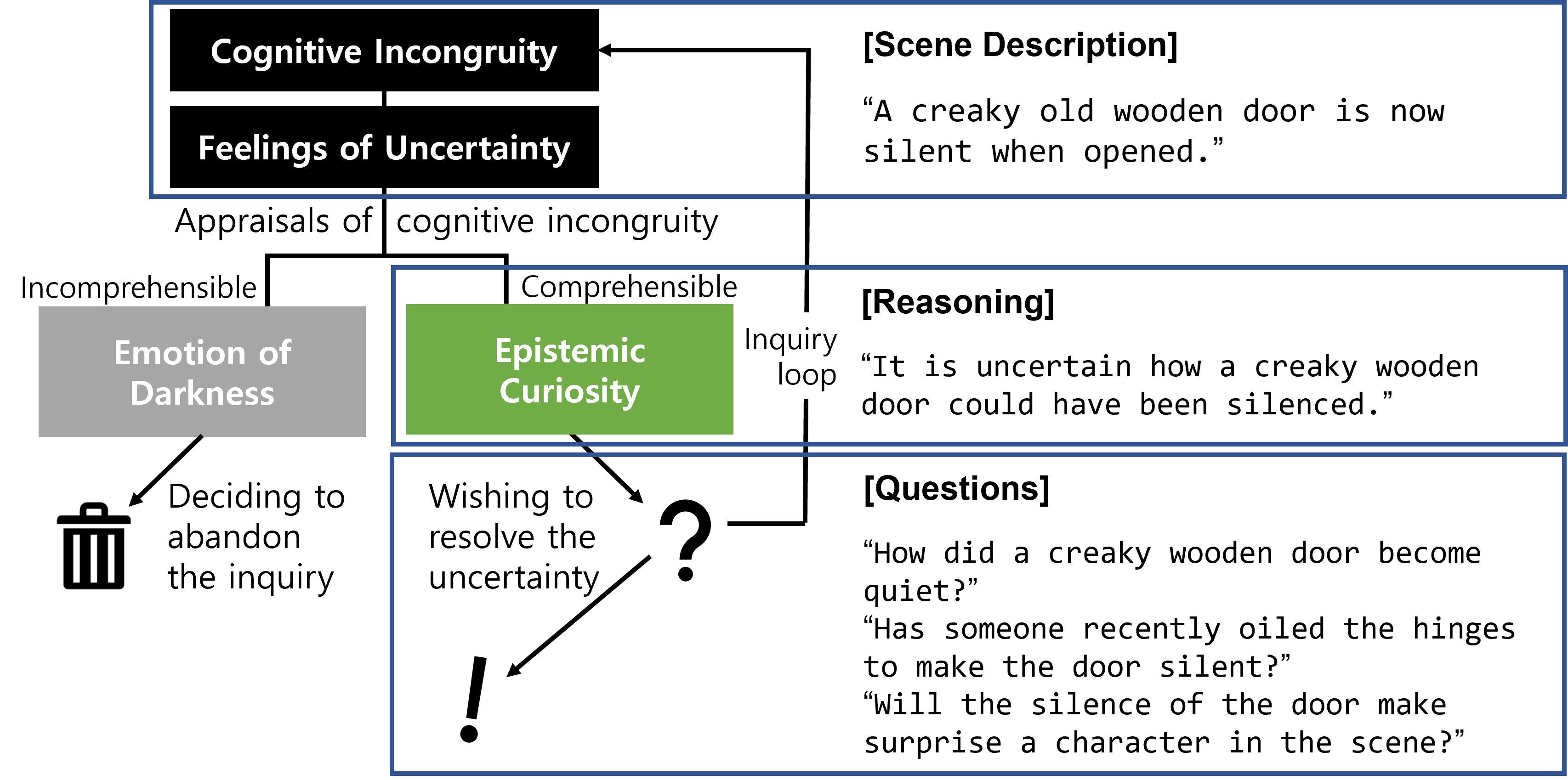}
\end{center}
\caption{The main concept of the CAUS dataset aligned with human cognition. The uncertain \texttt{Scene Description} provides a context that causes ``epistemic curiosity." The \texttt{Reasoning} sentences point out the uncertain point in a given scene. And the \texttt{Questions} represent efforts to resolve the uncertainty.} 
\label{concept}
\end{figure}

Based on the flow of the uncertainty resolution process in humans, as presented in Fig.~\ref{concept}, we suggest a dataset that aims to emulate epistemic curiosity. 
Briefly, when we encounter uncertainty, we first identify missing information and use a suitable thinking strategy, including making answerable questions.

\begin{table*}[t]
    \caption {Example of configuration for each uncertainty class in the dataset. For detailed explanations of K-type and Q-type, refer to the \textbf{Question Classification} section, and \cite{shin2023uncertainty}}
    \label {tab:ex}
    {\renewcommand{\arraystretch}{1.0}%
    \begin{tabular}{p{0.07\textwidth}|p{0.87\textwidth}}
    \hline
    Uncertain\\ty Class   & Contents  \\
    \hline
    Object & 
        \begin{tabular}[c]{@{}l@{}}
            \textbf{\{Scene Description\}}: A hand mirror is seen on the kitchen counter. \\ 
            \textbf{\{Reasoning\}}: It's unclear why a hand mirror would be found on the kitchen counter, a place typically\\ reserved for cooking utensils and ingredients.\\ 
                \begin{tabular}[c]{p{9.5cm}p{2.4cm}p{3cm}}
                \textbf{\{Questions\}}: &[K-type]&[Q-type]\\
                Why is a hand mirror on the kitchen counter? &Causality &Intention disclosure\\ 
                Who left the mirror there? &Identity &Concept completion\\
                Was it used for a specific purpose in the kitchen? &Intention &Verification \\
                Is it common for this mirror to be misplaced? &Attributes &Verification \\
                What is the usual location of this mirror? &Spatial layout&Concept completion
                 \end{tabular} \\
        \end{tabular} \\
    \hline
    Intention & 
        \begin{tabular}[c]{@{}l@{}}
            \textbf{\{Scene Description\}}: A classmate sits alone during lunch breaks and avoids social interactions. \\ 
            \textbf{\{Reasoning\}}: It's unclear why the classmate chooses to sit alone and avoid social interactions during\\ lunch breaks.\\
                \begin{tabular}[l]{p{9.5cm}p{2.5cm}p{3cm}}
                \textbf{\{Questions\}}: &[K-type]&[Q-type]\\
                Why does the classmate prefer solitude during lunch breaks? &Intention&Intention disclosure\\ 
                What is the classmate's general attitude towards social interactions? &Internal state&Judging\\
                Are there any observable factors contributing to the classmate's isolation? &Causality&Cause elucidation\\
                Has the classmate always behaved this way or is it a recent change? &Temporal relation&Case specification\\
                How do other classmates react to this individual's behavior? &Internal state&Result account
                 \end{tabular} \\
        \end{tabular} \\
    \hline
    Event &     
        \begin{tabular}[c]{@{}l@{}}
            \textbf{\{Scene Description\}}: The bottom-ranked team now holds the championship trophy. \\ 
            \textbf{\{Reasoning\}}: It's unclear how the bottom-ranked team, who were presumably underperforming, managed\\ to secure the championship trophy.\\
                \begin{tabular}[l]{p{9.5cm}p{2.4cm}p{3cm}}
                \textbf{\{Questions\}}: &[K-type]&[Q-type]\\
                How did the bottom-ranked team manage to win the championship? &Procedure&Method explication\\ 
                What strategies did they employ to overcome their ranking? &Procedure&Method explication\\
                Who were the key players in their victory? &Identity&Concept completion\\
                What was the reaction of the top-ranked teams? &Internal state&Result account\\
                How will this victory impact the team's future performance? &Causality&Expectation
                 \end{tabular} \\
        \end{tabular} \\                 
    \hline
    \end{tabular}}
\end{table*}

\begin{figure}[ht]
\begin{center}
\includegraphics[width=\linewidth]{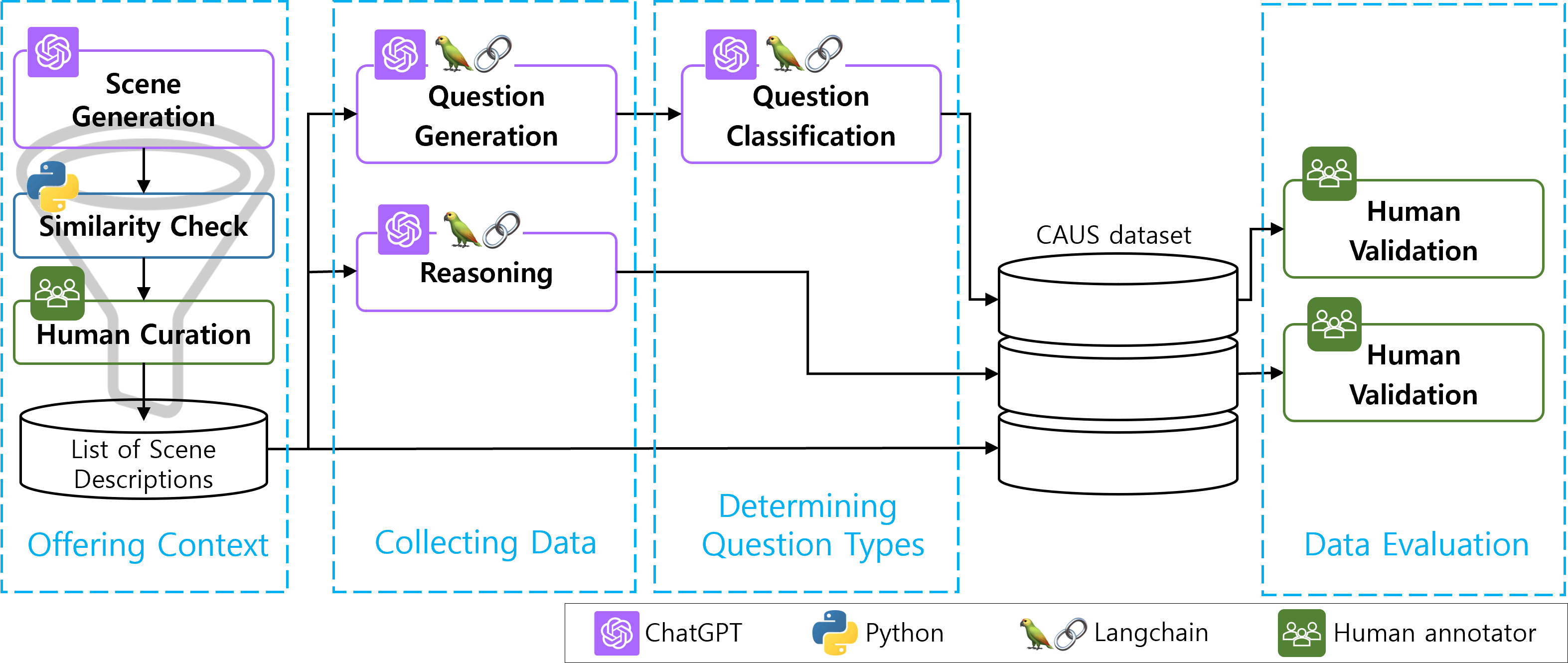}
\end{center} 
\caption{The pipeline for creating a dataset involves collecting, classifying, and evaluating potential questions evoked from scene descriptions. Symbols in the bottom box represent frameworks used in the pipeline.} 
\label{pipeline}
\end{figure}

The {CAUS} dataset consists of 1K of \textit{scene description} sentences, 1K of \textit{reasoning} sentences, and 5K of \textit{inquisitive question} sentences, which are all written in English. 
The data configuration is presented in Table~\ref{tab:ex}, and the entire process of building the dataset is illustrated in Fig.~\ref{pipeline}.

\subsection{Scene Description Generation}

We crafted scene description texts containing intentional inconsistencies. These inconsistencies were designed to create objective information gaps and cognitive disequilibrium without involving any social dynamics. The texts provide a context that elicits epistemic curiosity and encourages exploration beyond the surface level.

Three uncertainty classes were established within scenes for the following reasons: 1) To avoid any bias towards limited features in the scene generation, and 2) To validate the capability to appropriately interpret and respond to nuances of different types of uncertainty.

We first carefully inspected the Situation Model \cite{zwaan1999situation} and the Event-Indexing Model \cite{zwaan1995event} in order to refer to the well-formed structural framework for understanding narratives and examine the various elements they suggest (e.g., events, time, location, characters, objects, causality, motives, purposes, and plans). 
We then decided to focus on three distinct classes based on key elements of uncertainty: Object, Intention, and Event uncertainty. \textbf{Object uncertainty} refers to instances where objects in a scene are contextually out of place. 
\textbf{Intention uncertainty} addresses situations with unclear motivations behind a character's actions. \textbf{Event uncertainty} involves ambiguity in a specific stage of an event within the scene. This tripartite classification allows for a comprehensive elucidation of various uncertainties in scene interpretation.

We used the GPT-4 API released from OpenAI\footnote{https://openai.com/} to generate diverse scene descriptions. 
We provided the model with zero-shot instructions and set the temperature to 0.7 and 1, which is recommended for diverse outcomes \cite{openai2024temperature}.
After gaining excessive sentences, we checked for similar sentences using the cosine similarity algorithm, and two researchers reviewed them to remove duplicates.
The deduplicated list still required active human curation. 
Two researchers deleted or modified 1) inappropriate scenes involving biases on occupation, gender, etc., 2) unrealistic scenes that do not follow the laws of physics, 3) scenes in which uncertainty was diminished by subtext, and 4) scenes that lacked unexpectedness. The filtered list was then finalized by a third researcher to establish a list of scenes that contained uncertainty. After the meticulous inspection, we attained 1,000 scene description sentences, of which 328 were object-related, 364 were intention-related, and 308 were event-related uncertainties. 

\subsection{Reasoning and Query Generation}

\subsubsection{Reasoning} 
Humans appraise uncertainty at a metacognitive level during the initial phase of resolution process (Fig.~\ref{concept}). We implemented the core function of uncertainty resolution by generating sentences that point out unclear aspects of the given scene through inferring. For the inference process, we adopted the GPT-4-0613 model, which demonstrated the highest test performance, with a temperature setting of 0 to produce deterministic outcomes. 
The reasoning process is carried out with the main instruction: \texttt {"Point out something unclear or uncertain from the scene in one statement using a relation pronoun (who, what, where, when, how, or why)"}.

\begin{table}[t]
\begin{center}
\caption{K-type category} 
\label{tab:k} 
\vskip 0.05in
    \begin{tabular}{ll|ll} 
    \hline
    \# & Categories & \# & Categories\\
    \hline
    K1 & Identity & K7 & Contents  \\
    K2 & Class &  K8 & Procedure  \\
    K3 & Attributes & K9 & Causality  \\
    K4 & Quantities &  K10 & Intention  \\
    K5 & Spatial layout & K11 & Internal state  \\
    K6 & Temporal relation & & \\    
    \hline
\end{tabular} 
\end{center}
\end{table}

\subsubsection{Query Generation} 
To implement the essential efforts for uncertainty resolution, asking questions, we presented the scene description to the GPT-4-0613 model and instructed it to generate questions addressing these uncertain aspects.
As with the reasoning phase, we leveraged a model demonstrating the highest performance in tests.
We also set the temperature to 0 to generate deterministic outcomes.
To ensure a diverse range of questions, we instructed the model to sequentially create questions, starting from those \textit{spotting the most uncertain feature} to those \textit{exploring the situation}. 
In addition, no additional constraints were placed to allow us to observe the model's question-generating behavior. This approach aimed to capture a wide spectrum of inquiry types, reflecting the context of the scenes described.
The questioning process is carried out with the main instruction: \texttt {"Create five different terse questions that can be derived from a scene, from directly targeting the uncertain aspect to gathering additional information from the situation."}

\subsection{Question Classification}
Alongside the query generation, we conducted 2-dimensional classifications of the generated questions referring to our prior work \cite{shin2023uncertainty}. 
The first dimension is knowledge type (K-type), which identifies missing information that can be the source or target of inquiry. 
Table \ref{tab:k} shows eleven different K-type categories that specify the potential class of missing information in interactive situations.
These K-types range from simple and objective to complex and subjective as the number increases.

The second dimension of question classification is question type (Q-type), which represents the issue of how to express the inquiry. 
Table \ref{tab:q} displays fourteen different Q-type categories that classify inquiry expressions based on pragmatics.
As the number grows, the questions become more profound and subjective, like the K-type categories.

\begin{table}[t]
\caption{Q-type category} 
\label{tab:q} 
\begin{center}
\vskip 0.05in
    \begin{tabular}{ll|ll} 
    \hline
    \# & Categories & \# & Categories \\
    \hline
    Q1 & Verification & Q8 & Interpretation  \\
    Q2 & Case specification & Q9 & Cause elucidation \\
    Q3 & Concept completion & Q10 & Intention disclosure \\
    Q4 & Feature specification & Q11 & Result account \\
    Q5 & Quantification & Q12 & Method explication  \\
    Q6 & Definition & Q13 & Expectation  \\
    Q7 & Comparison & Q14 & Judging  \\
    \hline
\end{tabular} 
\end{center}
\end{table}

\begin{figure*}[t]
\begin{center}
\includegraphics[width=\linewidth]{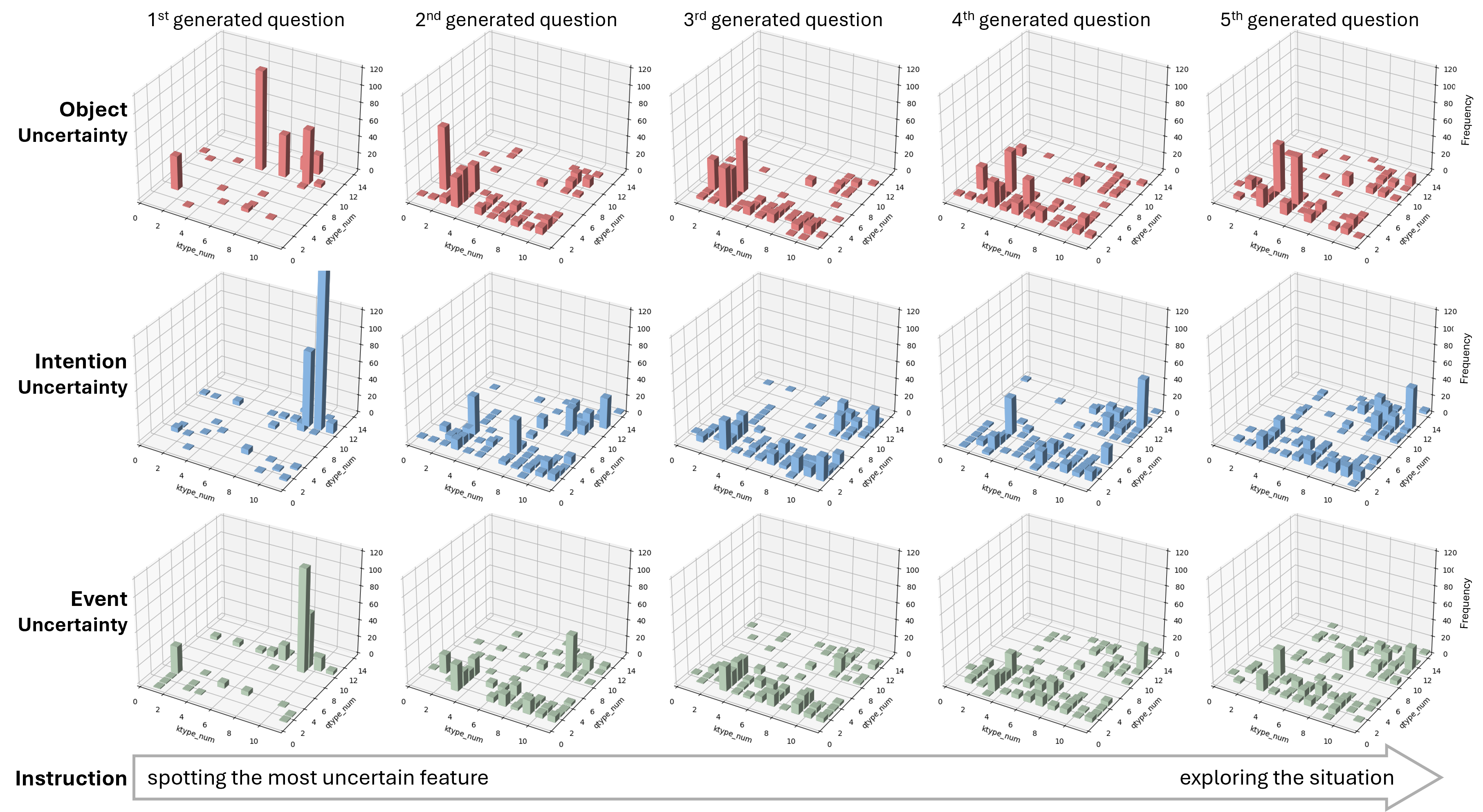}
\end{center}
\caption{Question classification results. Each row indicates the scene class (i.e., (Top)\textit{Object Uncertainty}. (Middle)\textit{Intention Uncertainty}. (Bottom){Event Uncertainty.}). Each column shows the generation order of the question. In each plot, the position of the bars corresponds to the question category, coordinated by k-type and q-type. The height of the bars indicates the frequency of the question for that type pair.}  
\label{freq}
\end{figure*}

\begin{figure}[ht]
\begin{center}
\includegraphics[width=0.6\columnwidth]{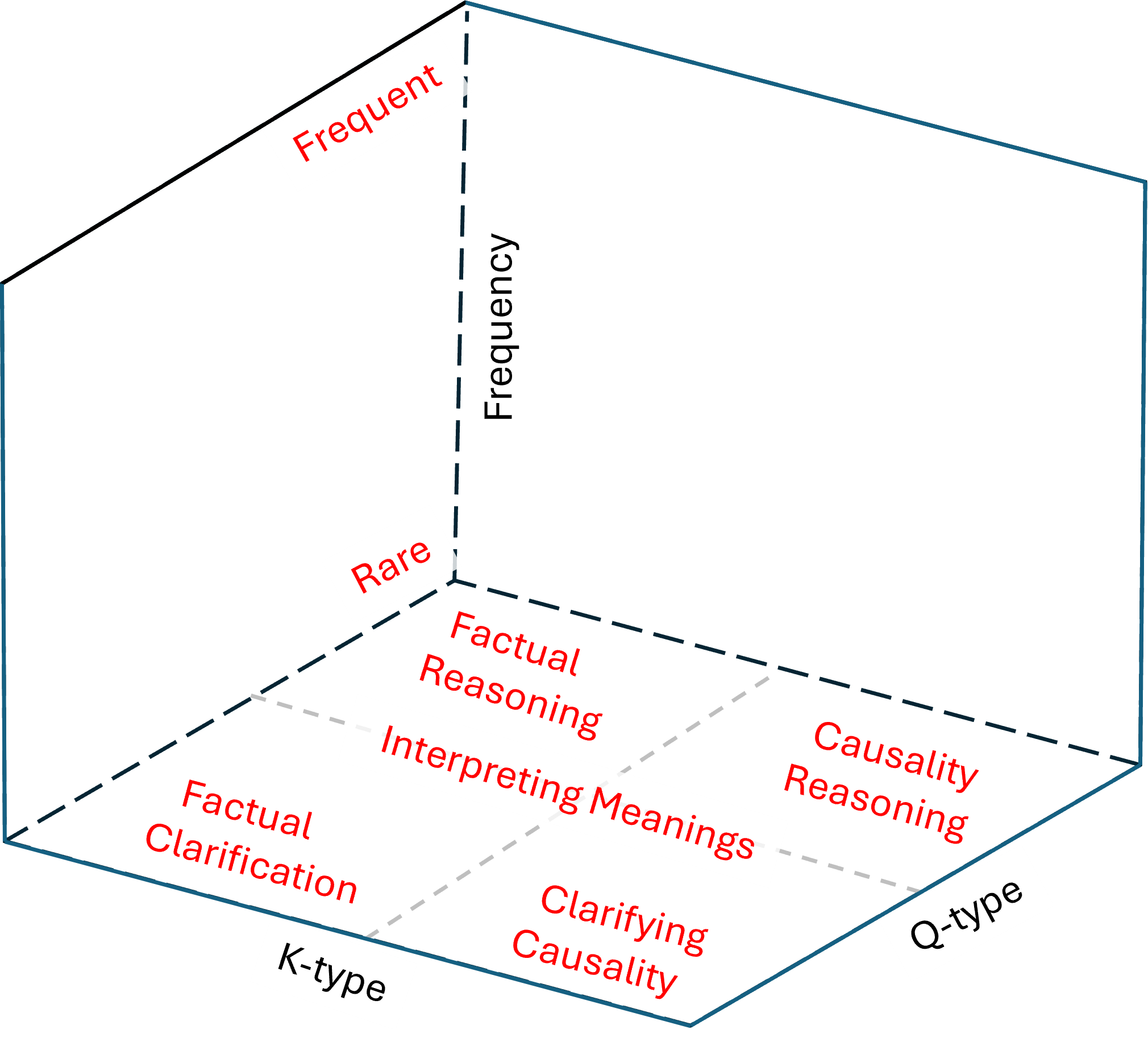}
\end{center}
\caption{Denoting the question space for displaying question classification}
\label{space}
\end{figure}

In question classification, we utilized the GPT-4-0613 model and the Langchain library\footnote{An open-source Python package that offers a most standardized interface for various LLM applications compatible with experimenting with different ideas, prompts, and models. https://python.langchain.com/} by providing detailed instruction prompts for categorizing questions into different types.
Although we fixed the model's temperature to 0 to achieve deterministic outcomes, slight variations were observed in the classification results with each iteration due to the inherent nature of LLMs. 
To mitigate this problem, we repeated the process three times and adopted the model's classification if at least two out of three repetitions agreed. 
Despite having more than ten category options for both K-type and Q-type, the model showed consistent classification performance.
Across all three iterations, a majority of questions (90.6\% for K-type, 91.4\% for Q-type) were categorized identically, and most of the remaining questions (9.1\% for K-type, 8.3\% for Q-type) were made two same results out of three iterations. 
In very rare cases ($\sim$ 0.3\%) where all three outcomes differed, we labeled the outcome `0, Undetermined.'
The classification results collected from each condition were displayed in a question space (Fig.~\ref{freq}, further detailed in the \textbf{Experimental Results} section).  
The meaning of each position within the question space is summarized in Fig.~\ref{space}.

\subsection{Evaluation}
\subsubsection{Reasoning} 
A random sample of 100 inferences, representing 10\% of the dataset, was taken. Two researchers reviewed the generated inference sentences separately to determine whether they accurately pointed to the uncertainty in the scene. Then, the results of the reviews were consolidated and discussed to assess accuracy.

\subsubsection{Question Classification} 
We evaluated the classification results of 500 questions from 100 randomly selected scene descriptions leveraging the GPT model. The 500 questions represent 10\% of the dataset. Two researchers created human-generated ground truth according to the definition of K-type and Q-type criteria. In creating the ground truth, researchers achieved an 85\% inter-rater agreement, and discrepancies were resolved through discussion.

\section {Experimental Results}

\subsection{Question Classification Results}

Fig.~\ref{freq} shows the distribution of queries, categorizing questions into K-type and Q-type. Five questions were generated for each scene description sentence, labeled with K-type and Q-type criteria, and plotted in the question spaces according to their labels. The plots were organized according to the generation order of each sentence.

The leftmost column shows classification result of the first generated questions, resulting in a highly clustered pattern. This concentration toward specific types reflects the instruction intended to spot the uncertain aspect directly. 
Most clusters fall under the \textit{Causality Reasoning} area, focusing on inquiring about the antecedents or intentions behind the event. At the same time, a few pertain to \textit{Factual Clarification}, focusing on identifying the subject of the action.
However, in intention uncertainty (blue bars in the middle row), it is observed that almost all questions fall within the \textit{Causality Reasoning} region, reflecting contexts where verifying the subject of the action is unnecessary.

On the other hand, those sets toward the right are more diverse and spread evenly throughout the question space, reflecting the instruction to gather additional information from the situation.
In the object uncertainty (red bars in the top row), later questions predominantly belong to the \textit{factual clarification} category, focusing on the attributes of objects. 
Conversely, for the intention uncertainty (blue bars in the middle row), the questions commonly aim to infer or clarify the causes or effects of actions. 
Regarding event uncertainty (green bars in the bottom row), a blend of object-related and behavior-related uncertainties, there is a tendency for the aforementioned patterns to intermingle, reflecting a combination of both attributes and causality.

\subsection{Evaluation Results}
\subsubsection{Evaluation on Reasoning Sentences}
Upon evaluating a randomly sampled set of 100 inference sentences, it was determined that over 90\% of the sentences were deemed appropriate. Sixty-one sentences were inferred within the scope of the words presented in the scene (e.g., ``It's unclear why the colleague always keeps their office door closed."), while thirty sentences incorporated contextual cues for inference (e.g., ``It's unclear why a renowned painter, \textit{who typically exhibits in galleries or museums}, is selling his pieces on the street."). However, a minority of nine sentences constituted inappropriate inferences. These inappropriate inferences distorted the content by introducing irrelevant clues or departed from the laws of physics (e.g., ``It's unclear how a refrigerator magnet is attached to a car door, \textit{which is typically made of materials not receptive to magnets}.").

\subsubsection{Evaluation on Question Classification}
We recorded the number of matches with the human-generated ground truth to evaluate the model-generated question classification.
There was an 83.2\% match (416 questions) between the model and human ground truth for the K-type questions. Similarly, for Q-type questions, an 83.6\% match (418 questions) was observed. We noted that items yielding different outcomes in all three iterations were challenging even for human evaluators to categorize into a single type. Additionally, in cases where two out of three iterations produced the same result, the differing third outcome tended to be semantically adjacent to the other types.

\section {Discussion}
In this study, we designed a novel dataset based on human cognition, along with a pipeline generating human-like questions and question classification structures. Our approach leveraged cutting-edge tools such as the GPT-4 model, and we performed in-depth evaluations to assess its performance. 
We made the controlled uncertainty with scene descriptions that offer a text version of Out Of Distribution (OOD). The uncertainty was fine-tuned to remain predictable, though not as tightly structured as in a game context. The main focus was understanding the LLM's capability to identify and inquire about uncertain elements within a given context. We also explored whether LLM can capture the nuanced content and format of given questions by the question classification.  

To wrap up, we revisit the two questions posed at the outset.
Firstly, we asked: Are LLMs inherently incompetent at asking questions? Our findings challenge this prevailing idea, demonstrating that LLMs can generate appropriate questions with proper context and instruction. Moreover, LLM's high question classification performance also revealed that LLM is good at predicting the properties of questions, as if the model understands the questioner's purpose and motivation. 
In conclusion, LLMs seem to have the potential to ask and grasp the nuance. 

Secondly, we considered: Can AI models be improved by implementing human questioning strategies when dealing with uncertainty? While this specific question was not directly addressed in our study, the findings offer a promising outlook on the potential applicability of such strategies. As motivated by the Vicuna model, which effectively enhanced the capabilities of a 13 billion parameter small LLM through fine-tuning for multi-round and long conversations \cite{vicuna2023}, it is evident that the strategic application of appropriate datasets can significantly bolster the utility of LLMs for specific purposes.

Acknowledging the limitations of our research, it is important to note that our work is highly sensitive to the specific prompting used. 
Our procedure, while optimized for the GPT-4 model, may not yield consistent results in the question classification phase with different models.
Currently, this exploratory approach, aiming for optimal outcomes, is commonly seen in other studies involving LLMs \cite{Reynolds2021Prompt, webson2022prompt}.
However, this sensitivity underscores the necessity for developing more robust and model-agnostic prompting techniques.
We also acknowledge an aspect in our hypothetical work that remains unaddressed, yet is crucial in human questioning: the role of desires and social interactions. 
This omission points to the need for further research integrating these human factors into the study of question generation and its application.

In summary, our research suggests that LLMs are not fundamentally limited in their ability to ask questions, and the application of human-like questioning strategies in AI models, particularly in dealing with uncertainties, holds substantial promise for future advancements.


\section{Acknowledgments}
We deeply thank the reviewers for providing kind and helpful comments. And this work was supported by Institute of Information \& communications Technology Planning \& Evaluation (IITP) grant funded by the Korea government(MSIT) (No. 20220-00951, Development of Uncertainty-Aware Agents Learning by Asking Questions)

\bibliographystyle{apacite}

\setlength{\bibleftmargin}{.125in}
\setlength{\bibindent}{-\bibleftmargin}

\bibliography{cogsci_CAUS}

\begin{thebibliography}{}

\bibitem [\protect \citeauthoryear {%
Brown%
\ \protect \BOthers {.}}{%
Brown%
\ \protect \BOthers {.}}{%
{\protect \APACyear {2020}}%
}]{%
brown2020language}
\APACinsertmetastar {%
brown2020language}%
\begin{APACrefauthors}%
Brown, T.%
, Mann, B.%
, Ryder, N.%
, Subbiah, M.%
, Kaplan, J\BPBI D.%
, Dhariwal, P.%
\BDBL {}others%
\end{APACrefauthors}%
\unskip\
\newblock
\APACrefYearMonthDay{2020}{}{}.
\newblock
{\BBOQ}\APACrefatitle {Language models are few-shot learners} {Language models are few-shot learners}.{\BBCQ}
\newblock
\APACjournalVolNumPages{Advances in neural information processing systems}{33}{}{1877--1901}.
\PrintBackRefs{\CurrentBib}

\bibitem [\protect \citeauthoryear {%
Cervera%
, Wang%
\BCBL {}\ \BBA {} Hayden%
}{%
Cervera%
\ \protect \BOthers {.}}{%
{\protect \APACyear {2020}}%
}]{%
Cervera2020SysNeuro}
\APACinsertmetastar {%
Cervera2020SysNeuro}%
\begin{APACrefauthors}%
Cervera, R\BPBI L.%
, Wang, M\BPBI Z.%
\BCBL {}\ \BBA {} Hayden, B\BPBI Y.%
\end{APACrefauthors}%
\unskip\
\newblock
\APACrefYearMonthDay{2020}{}{}.
\newblock
{\BBOQ}\APACrefatitle {Systems neuroscience of curiosity} {Systems neuroscience of curiosity}.{\BBCQ}
\newblock
\APACjournalVolNumPages{Current Opinion in Behavioral Sciences}{35}{}{48--55}.
\newblock
\begin{APACrefDOI} \doi{10.1016/j.cobeha.2020.06.011} \end{APACrefDOI}
\PrintBackRefs{\CurrentBib}

\bibitem [\protect \citeauthoryear {%
Chen%
, Yang%
, Hauff%
\BCBL {}\ \BBA {} Houben%
}{%
Chen%
\ \protect \BOthers {.}}{%
{\protect \APACyear {2018}}%
}]{%
chen2018learningq}
\APACinsertmetastar {%
chen2018learningq}%
\begin{APACrefauthors}%
Chen, G.%
, Yang, J.%
, Hauff, C.%
\BCBL {}\ \BBA {} Houben, G\BHBI J.%
\end{APACrefauthors}%
\unskip\
\newblock
\APACrefYearMonthDay{2018}{}{}.
\newblock
{\BBOQ}\APACrefatitle {LearningQ: a large-scale dataset for educational question generation} {Learningq: a large-scale dataset for educational question generation}.{\BBCQ}
\newblock
\BIn{} \APACrefbtitle {Proceedings of the international AAAI conference on web and social media} {Proceedings of the international aaai conference on web and social media}\ (\BVOL~12).
\newblock
\begin{APACrefDOI} \doi{10.1609/icwsm.v12i1.14987} \end{APACrefDOI}
\PrintBackRefs{\CurrentBib}

\bibitem [\protect \citeauthoryear {%
Chiang%
\ \protect \BOthers {.}}{%
Chiang%
\ \protect \BOthers {.}}{%
{\protect \APACyear {2023}}%
}]{%
vicuna2023}
\APACinsertmetastar {%
vicuna2023}%
\begin{APACrefauthors}%
Chiang, W\BHBI L.%
, Li, Z.%
, Lin, Z.%
, Sheng, Y.%
, Wu, Z.%
, Zhang, H.%
\BDBL {}Xing, E\BPBI P.%
\end{APACrefauthors}%
\unskip\
\newblock
\APACrefYearMonthDay{2023}{March}{}.
\newblock
\APACrefbtitle {Vicuna: An Open-Source Chatbot Impressing GPT-4 with 90\%* ChatGPT Quality.} {Vicuna: An open-source chatbot impressing gpt-4 with 90\%* chatgpt quality.}
\newblock
\begin{APACrefURL} \url{https://lmsys.org/blog/2023-03-30-vicuna/} \end{APACrefURL}
\newblock
\APACrefnote{Last accessed: 2024-05-09}
\PrintBackRefs{\CurrentBib}

\bibitem [\protect \citeauthoryear {%
Chouinard%
, Harris%
\BCBL {}\ \BBA {} Maratsos%
}{%
Chouinard%
\ \protect \BOthers {.}}{%
{\protect \APACyear {2007}}%
}]{%
chouinard2007children}
\APACinsertmetastar {%
chouinard2007children}%
\begin{APACrefauthors}%
Chouinard, M\BPBI M.%
, Harris, P\BPBI L.%
\BCBL {}\ \BBA {} Maratsos, M\BPBI P.%
\end{APACrefauthors}%
\unskip\
\newblock
\APACrefYearMonthDay{2007}{}{}.
\newblock
{\BBOQ}\APACrefatitle {Children's questions: A mechanism for cognitive development} {Children's questions: A mechanism for cognitive development}.{\BBCQ}
\newblock
\APACjournalVolNumPages{Monographs of the society for research in child development}{}{}{i--129}.
\PrintBackRefs{\CurrentBib}

\bibitem [\protect \citeauthoryear {%
Damassino%
}{%
Damassino%
}{%
{\protect \APACyear {2020}}%
}]{%
damassino2020QTT}
\APACinsertmetastar {%
damassino2020QTT}%
\begin{APACrefauthors}%
Damassino, N.%
\end{APACrefauthors}%
\unskip\
\newblock
\APACrefYearMonthDay{2020}{}{}.
\newblock
{\BBOQ}\APACrefatitle {The questioning turing test} {The questioning turing test}.{\BBCQ}
\newblock
\APACjournalVolNumPages{Minds and Machines}{30}{4}{563--587}.
\PrintBackRefs{\CurrentBib}

\bibitem [\protect \citeauthoryear {%
Duan%
, Tang%
, Chen%
\BCBL {}\ \BBA {} Zhou%
}{%
Duan%
\ \protect \BOthers {.}}{%
{\protect \APACyear {2017}}%
}]{%
duan2017question}
\APACinsertmetastar {%
duan2017question}%
\begin{APACrefauthors}%
Duan, N.%
, Tang, D.%
, Chen, P.%
\BCBL {}\ \BBA {} Zhou, M.%
\end{APACrefauthors}%
\unskip\
\newblock
\APACrefYearMonthDay{2017}{}{}.
\newblock
{\BBOQ}\APACrefatitle {Question generation for question answering} {Question generation for question answering}.{\BBCQ}
\newblock
\BIn{} \APACrefbtitle {Proceedings of the 2017 conference on empirical methods in natural language processing} {Proceedings of the 2017 conference on empirical methods in natural language processing}\ (\BPGS\ 866--874).
\newblock
\begin{APACrefDOI} \doi{10.18653/v1/D17-1090} \end{APACrefDOI}
\PrintBackRefs{\CurrentBib}

\bibitem [\protect \citeauthoryear {%
Flammer%
}{%
Flammer%
}{%
{\protect \APACyear {1981}}%
}]{%
flammer1981towards}
\APACinsertmetastar {%
flammer1981towards}%
\begin{APACrefauthors}%
Flammer, A.%
\end{APACrefauthors}%
\unskip\
\newblock
\APACrefYearMonthDay{1981}{}{}.
\newblock
{\BBOQ}\APACrefatitle {Towards a theory of question asking} {Towards a theory of question asking}.{\BBCQ}
\newblock
\APACjournalVolNumPages{Psychological Research}{43}{4}{407--420}.
\PrintBackRefs{\CurrentBib}

\bibitem [\protect \citeauthoryear {%
Gallegos%
\ \protect \BOthers {.}}{%
Gallegos%
\ \protect \BOthers {.}}{%
{\protect \APACyear {2023}}%
}]{%
gallegos2023bias}
\APACinsertmetastar {%
gallegos2023bias}%
\begin{APACrefauthors}%
Gallegos, I\BPBI O.%
, Rossi, R\BPBI A.%
, Barrow, J.%
, Tanjim, M\BPBI M.%
, Kim, S.%
, Dernoncourt, F.%
\BDBL {}Ahmed, N\BPBI K.%
\end{APACrefauthors}%
\unskip\
\newblock
\APACrefYearMonthDay{2023}{}{}.
\newblock
{\BBOQ}\APACrefatitle {Bias and fairness in large language models: A survey} {Bias and fairness in large language models: A survey}.{\BBCQ}
\newblock
\APACjournalVolNumPages{arXiv preprint arXiv:2309.00770}{}{}{}.
\PrintBackRefs{\CurrentBib}

\bibitem [\protect \citeauthoryear {%
Golman%
\ \BBA {} Loewenstein%
}{%
Golman%
\ \BBA {} Loewenstein%
}{%
{\protect \APACyear {2018}}%
}]{%
golman2018desire}
\APACinsertmetastar {%
golman2018desire}%
\begin{APACrefauthors}%
Golman, R.%
\BCBT {}\ \BBA {} Loewenstein, G.%
\end{APACrefauthors}%
\unskip\
\newblock
\APACrefYearMonthDay{2018}{}{}.
\newblock
{\BBOQ}\APACrefatitle {The Desire for Knowledge and Wisdom} {The desire for knowledge and wisdom}.{\BBCQ}
\newblock
\BIn{} G.~Gordon\ (\BED), \APACrefbtitle {The New Science of Curiosity} {The new science of curiosity}\ (\BPGS\ 37--42).
\newblock
\APACaddressPublisher{Hauppauge, NY}{Nova Science Publishers, Inc.}
\PrintBackRefs{\CurrentBib}

\bibitem [\protect \citeauthoryear {%
Gong%
, Pan%
\BCBL {}\ \BBA {} Hu%
}{%
Gong%
\ \protect \BOthers {.}}{%
{\protect \APACyear {2022}}%
}]{%
gong2022khanq}
\APACinsertmetastar {%
gong2022khanq}%
\begin{APACrefauthors}%
Gong, H.%
, Pan, L.%
\BCBL {}\ \BBA {} Hu, H.%
\end{APACrefauthors}%
\unskip\
\newblock
\APACrefYearMonthDay{2022}{}{}.
\newblock
{\BBOQ}\APACrefatitle {Khanq: A dataset for generating deep questions in education} {Khanq: A dataset for generating deep questions in education}.{\BBCQ}
\newblock
\BIn{} \APACrefbtitle {Proceedings of the 29th International Conference on Computational Linguistics} {Proceedings of the 29th international conference on computational linguistics}\ (\BPGS\ 5925--5938).
\PrintBackRefs{\CurrentBib}

\bibitem [\protect \citeauthoryear {%
A.~Graesser%
, Ozuru%
\BCBL {}\ \BBA {} Sullins%
}{%
A.~Graesser%
\ \protect \BOthers {.}}{%
{\protect \APACyear {2009}}%
}]{%
Graesser2009whatGQ}
\APACinsertmetastar {%
Graesser2009whatGQ}%
\begin{APACrefauthors}%
Graesser, A.%
, Ozuru, Y.%
\BCBL {}\ \BBA {} Sullins, J.%
\end{APACrefauthors}%
\unskip\
\newblock
\APACrefYearMonthDay{2009}{}{}.
\newblock
{\BBOQ}\APACrefatitle {What is a good question?} {What is a good question?}{\BBCQ}
\newblock
\BIn{} M\BPBI G.~McKeown\ \BBA {} L.~Kucan\ (\BEDS), \APACrefbtitle {Bringing reading research to life} {Bringing reading research to life}\ (\BPGS\ 170--193).
\newblock
\APACaddressPublisher{New York, NY}{Guilford Press}.
\PrintBackRefs{\CurrentBib}

\bibitem [\protect \citeauthoryear {%
A\BPBI C.~Graesser%
\ \BBA {} Black%
}{%
A\BPBI C.~Graesser%
\ \BBA {} Black%
}{%
{\protect \APACyear {1985}}%
}]{%
graesser1985psychology}
\APACinsertmetastar {%
graesser1985psychology}%
\begin{APACrefauthors}%
Graesser, A\BPBI C.%
\BCBT {}\ \BBA {} Black, J\BPBI B.%
\end{APACrefauthors}%
\ (\BEDS).
\unskip\
\newblock
\APACrefYear{1985}.
\newblock
\APACrefbtitle {The Psychology of Questions} {The psychology of questions}.
\newblock
\APACaddressPublisher{}{Lawrence Erlbaum Associates, Inc.}
\PrintBackRefs{\CurrentBib}

\bibitem [\protect \citeauthoryear {%
A\BPBI C.~Graesser%
\ \BBA {} McMahen%
}{%
A\BPBI C.~Graesser%
\ \BBA {} McMahen%
}{%
{\protect \APACyear {1993}}%
}]{%
graesser1993anomalous}
\APACinsertmetastar {%
graesser1993anomalous}%
\begin{APACrefauthors}%
Graesser, A\BPBI C.%
\BCBT {}\ \BBA {} McMahen, C\BPBI L.%
\end{APACrefauthors}%
\unskip\
\newblock
\APACrefYearMonthDay{1993}{}{}.
\newblock
{\BBOQ}\APACrefatitle {Anomalous information triggers questions when adults solve quantitative problems and comprehend stories.} {Anomalous information triggers questions when adults solve quantitative problems and comprehend stories.}{\BBCQ}
\newblock
\APACjournalVolNumPages{Journal of Educational Psychology}{85}{1}{136}.
\PrintBackRefs{\CurrentBib}

\bibitem [\protect \citeauthoryear {%
A\BPBI C.~Graesser%
\ \BBA {} Olde%
}{%
A\BPBI C.~Graesser%
\ \BBA {} Olde%
}{%
{\protect \APACyear {2003}}%
}]{%
graesser2003does}
\APACinsertmetastar {%
graesser2003does}%
\begin{APACrefauthors}%
Graesser, A\BPBI C.%
\BCBT {}\ \BBA {} Olde, B\BPBI A.%
\end{APACrefauthors}%
\unskip\
\newblock
\APACrefYearMonthDay{2003}{}{}.
\newblock
{\BBOQ}\APACrefatitle {How does one know whether a person understands a device? The quality of the questions the person asks when the device breaks down.} {How does one know whether a person understands a device? the quality of the questions the person asks when the device breaks down.}{\BBCQ}
\newblock
\APACjournalVolNumPages{Journal of Educational Psychology}{95}{3}{524}.
\PrintBackRefs{\CurrentBib}

\bibitem [\protect \citeauthoryear {%
A\BPBI C.~Graesser%
, Person%
\BCBL {}\ \BBA {} Huber%
}{%
A\BPBI C.~Graesser%
\ \protect \BOthers {.}}{%
{\protect \APACyear {2013}}%
}]{%
graesser2013mechanisms}
\APACinsertmetastar {%
graesser2013mechanisms}%
\begin{APACrefauthors}%
Graesser, A\BPBI C.%
, Person, N.%
\BCBL {}\ \BBA {} Huber, J.%
\end{APACrefauthors}%
\unskip\
\newblock
\APACrefYearMonthDay{2013}{}{}.
\newblock
{\BBOQ}\APACrefatitle {Mechanisms that generate questions} {Mechanisms that generate questions}.{\BBCQ}
\newblock
\BIn{} \APACrefbtitle {Questions and information systems} {Questions and information systems}\ (\BPGS\ 167--188).
\newblock
\APACaddressPublisher{}{Psychology Press}.
\PrintBackRefs{\CurrentBib}

\bibitem [\protect \citeauthoryear {%
Huang%
, Yeomans%
, Brooks%
, Minson%
\BCBL {}\ \BBA {} Gino%
}{%
Huang%
\ \protect \BOthers {.}}{%
{\protect \APACyear {2017}}%
}]{%
huang2017doesn}
\APACinsertmetastar {%
huang2017doesn}%
\begin{APACrefauthors}%
Huang, K.%
, Yeomans, M.%
, Brooks, A\BPBI W.%
, Minson, J.%
\BCBL {}\ \BBA {} Gino, F.%
\end{APACrefauthors}%
\unskip\
\newblock
\APACrefYearMonthDay{2017}{}{}.
\newblock
{\BBOQ}\APACrefatitle {It doesn’t hurt to ask: Question-asking increases liking.} {It doesn’t hurt to ask: Question-asking increases liking.}{\BBCQ}
\newblock
\APACjournalVolNumPages{Journal of personality and social psychology}{113}{3}{430}.
\PrintBackRefs{\CurrentBib}

\bibitem [\protect \citeauthoryear {%
Jain%
, Zhang%
\BCBL {}\ \BBA {} Schwing%
}{%
Jain%
\ \protect \BOthers {.}}{%
{\protect \APACyear {2017}}%
}]{%
Jain2017CVPR}
\APACinsertmetastar {%
Jain2017CVPR}%
\begin{APACrefauthors}%
Jain, U.%
, Zhang, Z.%
\BCBL {}\ \BBA {} Schwing, A\BPBI G.%
\end{APACrefauthors}%
\unskip\
\newblock
\APACrefYearMonthDay{2017}{July}{}.
\newblock
{\BBOQ}\APACrefatitle {Creativity: Generating Diverse Questions Using Variational Autoencoders} {Creativity: Generating diverse questions using variational autoencoders}.{\BBCQ}
\newblock
\BIn{} \APACrefbtitle {Proceedings of the IEEE Conference on Computer Vision and Pattern Recognition (CVPR).} {Proceedings of the ieee conference on computer vision and pattern recognition (cvpr).}
\PrintBackRefs{\CurrentBib}

\bibitem [\protect \citeauthoryear {%
Krishna%
, Lee%
, Fei-Fei%
\BCBL {}\ \BBA {} Bernstein%
}{%
Krishna%
\ \protect \BOthers {.}}{%
{\protect \APACyear {2022}}%
}]{%
Krishna2022socialAI}
\APACinsertmetastar {%
Krishna2022socialAI}%
\begin{APACrefauthors}%
Krishna, R.%
, Lee, D.%
, Fei-Fei, L.%
\BCBL {}\ \BBA {} Bernstein, M\BPBI S.%
\end{APACrefauthors}%
\unskip\
\newblock
\APACrefYearMonthDay{2022}{}{}.
\newblock
{\BBOQ}\APACrefatitle {Socially situated artificial intelligence enables learning from human interaction} {Socially situated artificial intelligence enables learning from human interaction}.{\BBCQ}
\newblock
\APACjournalVolNumPages{Proceedings of the National Academy of Sciences}{119}{39}{e2115730119}.
\newblock
\begin{APACrefDOI} \doi{10.1073/pnas.2115730119} \end{APACrefDOI}
\PrintBackRefs{\CurrentBib}

\bibitem [\protect \citeauthoryear {%
Loewenstein%
}{%
Loewenstein%
}{%
{\protect \APACyear {1994}}%
}]{%
loewenstein1994psychology}
\APACinsertmetastar {%
loewenstein1994psychology}%
\begin{APACrefauthors}%
Loewenstein, G.%
\end{APACrefauthors}%
\unskip\
\newblock
\APACrefYearMonthDay{1994}{}{}.
\newblock
{\BBOQ}\APACrefatitle {The psychology of curiosity: A review and reinterpretation.} {The psychology of curiosity: A review and reinterpretation.}{\BBCQ}
\newblock
\APACjournalVolNumPages{Psychological bulletin}{116}{1}{75}.
\newblock
\begin{APACrefDOI} \doi{10.1037/0033-2909.116.1.75} \end{APACrefDOI}
\PrintBackRefs{\CurrentBib}

\bibitem [\protect \citeauthoryear {%
Macagno%
}{%
Macagno%
}{%
{\protect \APACyear {2023}}%
}]{%
macagno2023questions}
\APACinsertmetastar {%
macagno2023questions}%
\begin{APACrefauthors}%
Macagno, F.%
\end{APACrefauthors}%
\unskip\
\newblock
\APACrefYearMonthDay{2023}{}{}.
\newblock
{\BBOQ}\APACrefatitle {Questions as Dialogue Games. The Pragmatic Dimensions of “Authentic” Questions} {Questions as dialogue games. the pragmatic dimensions of “authentic” questions}.{\BBCQ}
\newblock
\APACjournalVolNumPages{Studies in Philosophy and Education}{42}{5}{519--539}.
\PrintBackRefs{\CurrentBib}

\bibitem [\protect \citeauthoryear {%
Misra%
\ \protect \BOthers {.}}{%
Misra%
\ \protect \BOthers {.}}{%
{\protect \APACyear {2018}}%
}]{%
misra2018learning}
\APACinsertmetastar {%
misra2018learning}%
\begin{APACrefauthors}%
Misra, I.%
, Girshick, R.%
, Fergus, R.%
, Hebert, M.%
, Gupta, A.%
\BCBL {}\ \BBA {} Van Der~Maaten, L.%
\end{APACrefauthors}%
\unskip\
\newblock
\APACrefYearMonthDay{2018}{}{}.
\newblock
{\BBOQ}\APACrefatitle {Learning by asking questions} {Learning by asking questions}.{\BBCQ}
\newblock
\BIn{} \APACrefbtitle {2018 IEEE/CVF Conference on Computer Vision and Pattern Recognition} {2018 ieee/cvf conference on computer vision and pattern recognition}\ (\BPGS\ 11--20).
\newblock
\APACaddressPublisher{Salt Lake City, UT, USA}{IEEE}.
\newblock
\begin{APACrefDOI} \doi{10.1109/CVPR.2018.00009} \end{APACrefDOI}
\PrintBackRefs{\CurrentBib}

\bibitem [\protect \citeauthoryear {%
Momennejad%
\ \protect \BOthers {.}}{%
Momennejad%
\ \protect \BOthers {.}}{%
{\protect \APACyear {2023}}%
}]{%
momennejad2023evaluating}
\APACinsertmetastar {%
momennejad2023evaluating}%
\begin{APACrefauthors}%
Momennejad, I.%
, Hasanbeig, H.%
, Frujeri, F\BPBI V.%
, Sharma, H.%
, Ness, R\BPBI O.%
, Jojic, N.%
\BDBL {}Larson, J.%
\end{APACrefauthors}%
\unskip\
\newblock
\APACrefYearMonthDay{2023}{}{}.
\newblock
{\BBOQ}\APACrefatitle {Evaluating cognitive maps in large language models with cogeval: No emergent planning} {Evaluating cognitive maps in large language models with cogeval: No emergent planning}.{\BBCQ}
\newblock
\APACjournalVolNumPages{Advances in neural information processing systems}{37}{}{}.
\PrintBackRefs{\CurrentBib}

\bibitem [\protect \citeauthoryear {%
{OpenAI Community}%
}{%
{OpenAI Community}%
}{%
{\protect \APACyear {2023}}%
}]{%
openai2024temperature}
\APACinsertmetastar {%
openai2024temperature}%
\begin{APACrefauthors}%
{OpenAI Community}.%
\end{APACrefauthors}%
\unskip\
\newblock
\APACrefYearMonthDay{2023}{April}{}.
\newblock
\APACrefbtitle {Cheat Sheet: Mastering Temperature and Top-p in ChatGPT API.} {Cheat sheet: Mastering temperature and top-p in chatgpt api.}
\newblock
\begin{APACrefURL} \url{https://community.openai.com/t/cheat-sheet- \allowbreak mastering-temperature-and-top-p-in-chatgpt-api/ \allowbreak 172683} \end{APACrefURL}
\newblock
\APACrefnote{Last accessed: 2024-05-09}
\PrintBackRefs{\CurrentBib}

\bibitem [\protect \citeauthoryear {%
Otero%
\ \BBA {} Graesser%
}{%
Otero%
\ \BBA {} Graesser%
}{%
{\protect \APACyear {2001}}%
}]{%
otero2001preg}
\APACinsertmetastar {%
otero2001preg}%
\begin{APACrefauthors}%
Otero, J.%
\BCBT {}\ \BBA {} Graesser, A\BPBI C.%
\end{APACrefauthors}%
\unskip\
\newblock
\APACrefYearMonthDay{2001}{}{}.
\newblock
{\BBOQ}\APACrefatitle {PREG: Elements of a model of question asking} {Preg: Elements of a model of question asking}.{\BBCQ}
\newblock
\APACjournalVolNumPages{Cognition and instruction}{19}{2}{143--175}.
\newblock
\begin{APACrefDOI} \doi{doi.org/10.1207/S1532690XCI1902\_01} \end{APACrefDOI}
\PrintBackRefs{\CurrentBib}

\bibitem [\protect \citeauthoryear {%
Reynolds%
\ \BBA {} McDonell%
}{%
Reynolds%
\ \BBA {} McDonell%
}{%
{\protect \APACyear {2021}}%
}]{%
Reynolds2021Prompt}
\APACinsertmetastar {%
Reynolds2021Prompt}%
\begin{APACrefauthors}%
Reynolds, L.%
\BCBT {}\ \BBA {} McDonell, K.%
\end{APACrefauthors}%
\unskip\
\newblock
\APACrefYearMonthDay{2021}{}{}.
\newblock
{\BBOQ}\APACrefatitle {Prompt Programming for Large Language Models: Beyond the Few-Shot Paradigm} {Prompt programming for large language models: Beyond the few-shot paradigm}.{\BBCQ}
\newblock
\BIn{} \APACrefbtitle {Extended Abstracts of the 2021 CHI Conference on Human Factors in Computing Systems.} {Extended abstracts of the 2021 chi conference on human factors in computing systems.}
\newblock
\APACaddressPublisher{New York, NY, USA}{Association for Computing Machinery}.
\newblock
\begin{APACrefDOI} \doi{10.1145/3411763.3451760} \end{APACrefDOI}
\PrintBackRefs{\CurrentBib}

\bibitem [\protect \citeauthoryear {%
Rosenshine%
, Meister%
\BCBL {}\ \BBA {} Chapman%
}{%
Rosenshine%
\ \protect \BOthers {.}}{%
{\protect \APACyear {1996}}%
}]{%
rosenshine1996teaching}
\APACinsertmetastar {%
rosenshine1996teaching}%
\begin{APACrefauthors}%
Rosenshine, B.%
, Meister, C.%
\BCBL {}\ \BBA {} Chapman, S.%
\end{APACrefauthors}%
\unskip\
\newblock
\APACrefYearMonthDay{1996}{}{}.
\newblock
{\BBOQ}\APACrefatitle {Teaching students to generate questions: A review of the intervention studies} {Teaching students to generate questions: A review of the intervention studies}.{\BBCQ}
\newblock
\APACjournalVolNumPages{Review of educational research}{66}{2}{181--221}.
\PrintBackRefs{\CurrentBib}

\bibitem [\protect \citeauthoryear {%
Rothe%
, Lake%
\BCBL {}\ \BBA {} Gureckis%
}{%
Rothe%
\ \protect \BOthers {.}}{%
{\protect \APACyear {2017}}%
}]{%
Rothe2017NIPS}
\APACinsertmetastar {%
Rothe2017NIPS}%
\begin{APACrefauthors}%
Rothe, A.%
, Lake, B\BPBI M.%
\BCBL {}\ \BBA {} Gureckis, T.%
\end{APACrefauthors}%
\unskip\
\newblock
\APACrefYearMonthDay{2017}{}{}.
\newblock
{\BBOQ}\APACrefatitle {Question Asking as Program Generation} {Question asking as program generation}.{\BBCQ}
\newblock
\BIn{} I.~Guyon\ \BOthers {.}\ (\BEDS), \APACrefbtitle {Advances in Neural Information Processing Systems} {Advances in neural information processing systems}\ (\BVOL~30).
\newblock
\APACaddressPublisher{}{Curran Associates, Inc.}
\PrintBackRefs{\CurrentBib}

\bibitem [\protect \citeauthoryear {%
Rothe%
, Lake%
\BCBL {}\ \BBA {} Gureckis%
}{%
Rothe%
\ \protect \BOthers {.}}{%
{\protect \APACyear {2018}}%
}]{%
rothe2018people}
\APACinsertmetastar {%
rothe2018people}%
\begin{APACrefauthors}%
Rothe, A.%
, Lake, B\BPBI M.%
\BCBL {}\ \BBA {} Gureckis, T\BPBI M.%
\end{APACrefauthors}%
\unskip\
\newblock
\APACrefYearMonthDay{2018}{}{}.
\newblock
{\BBOQ}\APACrefatitle {Do people ask good questions?} {Do people ask good questions?}{\BBCQ}
\newblock
\APACjournalVolNumPages{Computational Brain \& Behavior}{1}{1}{69--89}.
\newblock
\begin{APACrefDOI} \doi{10.1007/s42113-018-0005-5} \end{APACrefDOI}
\PrintBackRefs{\CurrentBib}

\bibitem [\protect \citeauthoryear {%
Shin%
, Jang%
, Cho%
\BCBL {}\ \BBA {} Ryu%
}{%
Shin%
\ \protect \BOthers {.}}{%
{\protect \APACyear {2023}}%
}]{%
shin2023uncertainty}
\APACinsertmetastar {%
shin2023uncertainty}%
\begin{APACrefauthors}%
Shin, M.%
, Jang, M.%
, Cho, M.%
\BCBL {}\ \BBA {} Ryu, J\BHBI K.%
\end{APACrefauthors}%
\unskip\
\newblock
\APACrefYearMonthDay{2023}{}{}.
\newblock
{\BBOQ}\APACrefatitle {Uncertainty-Resolving Questions for Social Robots} {Uncertainty-resolving questions for social robots}.{\BBCQ}
\newblock
\BIn{} \APACrefbtitle {Companion of the 2023 ACM/IEEE International Conference on Human-Robot Interaction} {Companion of the 2023 acm/ieee international conference on human-robot interaction}\ (\BPG~226–230).
\newblock
\APACaddressPublisher{New York, NY, USA}{Association for Computing Machinery}.
\newblock
\begin{APACrefDOI} \doi{10.1145/3568294.3580077} \end{APACrefDOI}
\PrintBackRefs{\CurrentBib}

\bibitem [\protect \citeauthoryear {%
Sultan%
, Chandel%
, Fernandez~Astudillo%
\BCBL {}\ \BBA {} Castelli%
}{%
Sultan%
\ \protect \BOthers {.}}{%
{\protect \APACyear {2020}}%
}]{%
sultan2020diverseQG}
\APACinsertmetastar {%
sultan2020diverseQG}%
\begin{APACrefauthors}%
Sultan, M\BPBI A.%
, Chandel, S.%
, Fernandez~Astudillo, R.%
\BCBL {}\ \BBA {} Castelli, V.%
\end{APACrefauthors}%
\unskip\
\newblock
\APACrefYearMonthDay{2020}{{\APACmonth{07}}}{}.
\newblock
{\BBOQ}\APACrefatitle {On the Importance of Diversity in Question Generation for {QA}} {On the importance of diversity in question generation for {QA}}.{\BBCQ}
\newblock
\BIn{} D.~Jurafsky, J.~Chai, N.~Schluter\BCBL {}\ \BBA {} J.~Tetreault\ (\BEDS), \APACrefbtitle {Proceedings of the 58th Annual Meeting of the Association for Computational Linguistics} {Proceedings of the 58th annual meeting of the association for computational linguistics}\ (\BPGS\ 5651--5656).
\newblock
\APACaddressPublisher{Online}{Association for Computational Linguistics}.
\newblock
\begin{APACrefDOI} \doi{10.18653/v1/2020.acl-main.500} \end{APACrefDOI}
\PrintBackRefs{\CurrentBib}

\bibitem [\protect \citeauthoryear {%
Toles%
, Huang%
, Yu%
\BCBL {}\ \BBA {} Gravano%
}{%
Toles%
\ \protect \BOthers {.}}{%
{\protect \APACyear {2023}}%
}]{%
toles2023good}
\APACinsertmetastar {%
toles2023good}%
\begin{APACrefauthors}%
Toles, M.%
, Huang, Y.%
, Yu, Z.%
\BCBL {}\ \BBA {} Gravano, L.%
\end{APACrefauthors}%
\unskip\
\newblock
\APACrefYearMonthDay{2023}{}{}.
\newblock
{\BBOQ}\APACrefatitle {What is a good question? Task-oriented asking with fact-level masking} {What is a good question? task-oriented asking with fact-level masking}.{\BBCQ}
\newblock
\APACjournalVolNumPages{arXiv preprint arXiv:2310.11571}{}{}{}.
\PrintBackRefs{\CurrentBib}

\bibitem [\protect \citeauthoryear {%
Vazard%
\ \BBA {} Audrin%
}{%
Vazard%
\ \BBA {} Audrin%
}{%
{\protect \APACyear {2022}}%
}]{%
Vazard2022noetic}
\APACinsertmetastar {%
Vazard2022noetic}%
\begin{APACrefauthors}%
Vazard, J.%
\BCBT {}\ \BBA {} Audrin, C.%
\end{APACrefauthors}%
\unskip\
\newblock
\APACrefYearMonthDay{2022}{}{}.
\newblock
{\BBOQ}\APACrefatitle {The noetic feeling of confusion} {The noetic feeling of confusion}.{\BBCQ}
\newblock
\APACjournalVolNumPages{Philosophical Psychology}{35}{5}{757-770}.
\newblock
\begin{APACrefDOI} \doi{10.1080/09515089.2021.2016675} \end{APACrefDOI}
\PrintBackRefs{\CurrentBib}

\bibitem [\protect \citeauthoryear {%
Wang%
, Yue%
\BCBL {}\ \BBA {} Sun%
}{%
Wang%
\ \protect \BOthers {.}}{%
{\protect \APACyear {2023}}%
}]{%
wang2023can}
\APACinsertmetastar {%
wang2023can}%
\begin{APACrefauthors}%
Wang, B.%
, Yue, X.%
\BCBL {}\ \BBA {} Sun, H.%
\end{APACrefauthors}%
\unskip\
\newblock
\APACrefYearMonthDay{2023}{}{}.
\newblock
{\BBOQ}\APACrefatitle {Can Chat{GPT} Defend its Belief in Truth? Evaluating {LLM} Reasoning via Debate} {Can chat{GPT} defend its belief in truth? evaluating {LLM} reasoning via debate}.{\BBCQ}
\newblock
\BIn{} \APACrefbtitle {The 2023 Conference on Empirical Methods in Natural Language Processing.} {The 2023 conference on empirical methods in natural language processing.}
\PrintBackRefs{\CurrentBib}

\bibitem [\protect \citeauthoryear {%
Webson%
\ \BBA {} Pavlick%
}{%
Webson%
\ \BBA {} Pavlick%
}{%
{\protect \APACyear {2022}}%
}]{%
webson2022prompt}
\APACinsertmetastar {%
webson2022prompt}%
\begin{APACrefauthors}%
Webson, A.%
\BCBT {}\ \BBA {} Pavlick, E.%
\end{APACrefauthors}%
\unskip\
\newblock
\APACrefYearMonthDay{2022}{{\APACmonth{07}}}{}.
\newblock
{\BBOQ}\APACrefatitle {Do Prompt-Based Models Really Understand the Meaning of Their Prompts?} {Do prompt-based models really understand the meaning of their prompts?}{\BBCQ}
\newblock
\BIn{} M.~Carpuat, M\BHBI C.~de Marneffe\BCBL {}\ \BBA {} I\BPBI V.~Meza~Ruiz\ (\BEDS), \APACrefbtitle {Proceedings of the 2022 Conference of the North American Chapter of the Association for Computational Linguistics: Human Language Technologies} {Proceedings of the 2022 conference of the north american chapter of the association for computational linguistics: Human language technologies}\ (\BPGS\ 2300--2344).
\newblock
\APACaddressPublisher{Seattle, United States}{Association for Computational Linguistics}.
\newblock
\begin{APACrefDOI} \doi{10.18653/v1/2022.naacl-main.167} \end{APACrefDOI}
\PrintBackRefs{\CurrentBib}

\bibitem [\protect \citeauthoryear {%
Wei%
\ \protect \BOthers {.}}{%
Wei%
\ \protect \BOthers {.}}{%
{\protect \APACyear {2022}}%
}]{%
wei2022emergent}
\APACinsertmetastar {%
wei2022emergent}%
\begin{APACrefauthors}%
Wei, J.%
, Tay, Y.%
, Bommasani, R.%
, Raffel, C.%
, Zoph, B.%
, Borgeaud, S.%
\BDBL {}Fedus, W.%
\end{APACrefauthors}%
\unskip\
\newblock
\APACrefYearMonthDay{2022}{}{}.
\newblock
{\BBOQ}\APACrefatitle {Emergent Abilities of Large Language Models} {Emergent abilities of large language models}.{\BBCQ}
\newblock
\APACjournalVolNumPages{Transactions on Machine Learning Research}{}{}{}.
\newblock
\APACrefnote{Survey Certification}
\PrintBackRefs{\CurrentBib}

\bibitem [\protect \citeauthoryear {%
West%
\ \protect \BOthers {.}}{%
West%
\ \protect \BOthers {.}}{%
{\protect \APACyear {2023}}%
}]{%
west2023generative}
\APACinsertmetastar {%
west2023generative}%
\begin{APACrefauthors}%
West, P.%
, Lu, X.%
, Dziri, N.%
, Brahman, F.%
, Li, L.%
, Hwang, J\BPBI D.%
\BDBL {}others%
\end{APACrefauthors}%
\unskip\
\newblock
\APACrefYearMonthDay{2023}{}{}.
\newblock
{\BBOQ}\APACrefatitle {THE GENERATIVE AI PARADOX:``What It Can Create, It May Not Understand”} {The generative ai paradox:``what it can create, it may not understand”}.{\BBCQ}
\newblock
\BIn{} \APACrefbtitle {The Twelfth International Conference on Learning Representations.} {The twelfth international conference on learning representations.}
\PrintBackRefs{\CurrentBib}

\bibitem [\protect \citeauthoryear {%
Yoon%
\ \BBA {} Bak%
}{%
Yoon%
\ \BBA {} Bak%
}{%
{\protect \APACyear {2023}}%
}]{%
yoon2023storybook}
\APACinsertmetastar {%
yoon2023storybook}%
\begin{APACrefauthors}%
Yoon, H.%
\BCBT {}\ \BBA {} Bak, J.%
\end{APACrefauthors}%
\unskip\
\newblock
\APACrefYearMonthDay{2023}{{\APACmonth{12}}}{}.
\newblock
{\BBOQ}\APACrefatitle {Diversity Enhanced Narrative Question Generation for Storybooks} {Diversity enhanced narrative question generation for storybooks}.{\BBCQ}
\newblock
\BIn{} H.~Bouamor, J.~Pino\BCBL {}\ \BBA {} K.~Bali\ (\BEDS), \APACrefbtitle {Proceedings of the 2023 Conference on Empirical Methods in Natural Language Processing} {Proceedings of the 2023 conference on empirical methods in natural language processing}\ (\BPGS\ 465--482).
\newblock
\APACaddressPublisher{Singapore}{Association for Computational Linguistics}.
\newblock
\begin{APACrefDOI} \doi{10.18653/v1/2023.emnlp-main.31} \end{APACrefDOI}
\PrintBackRefs{\CurrentBib}

\bibitem [\protect \citeauthoryear {%
Zhou%
, Zhang%
\BCBL {}\ \BBA {} Wu%
}{%
Zhou%
\ \protect \BOthers {.}}{%
{\protect \APACyear {2019}}%
}]{%
zhou2019question}
\APACinsertmetastar {%
zhou2019question}%
\begin{APACrefauthors}%
Zhou, W.%
, Zhang, M.%
\BCBL {}\ \BBA {} Wu, Y.%
\end{APACrefauthors}%
\unskip\
\newblock
\APACrefYearMonthDay{2019}{}{}.
\newblock
{\BBOQ}\APACrefatitle {Question-type Driven Question Generation} {Question-type driven question generation}.{\BBCQ}
\newblock
\BIn{} \APACrefbtitle {Proceedings of the 2019 Conference on Empirical Methods in Natural Language Processing and the 9th International Joint Conference on Natural Language Processing (EMNLP-IJCNLP)} {Proceedings of the 2019 conference on empirical methods in natural language processing and the 9th international joint conference on natural language processing (emnlp-ijcnlp)}\ (\BPGS\ 6032--6037).
\newblock
\begin{APACrefDOI} \doi{10.18653/v1/D19-1622} \end{APACrefDOI}
\PrintBackRefs{\CurrentBib}

\bibitem [\protect \citeauthoryear {%
Zwaan%
}{%
Zwaan%
}{%
{\protect \APACyear {1999}}%
}]{%
zwaan1999situation}
\APACinsertmetastar {%
zwaan1999situation}%
\begin{APACrefauthors}%
Zwaan, R\BPBI A.%
\end{APACrefauthors}%
\unskip\
\newblock
\APACrefYearMonthDay{1999}{}{}.
\newblock
{\BBOQ}\APACrefatitle {Situation models: The mental leap into imagined worlds} {Situation models: The mental leap into imagined worlds}.{\BBCQ}
\newblock
\APACjournalVolNumPages{Current directions in psychological science}{8}{1}{15--18}.
\PrintBackRefs{\CurrentBib}

\bibitem [\protect \citeauthoryear {%
Zwaan%
, Langston%
\BCBL {}\ \BBA {} Graesser%
}{%
Zwaan%
\ \protect \BOthers {.}}{%
{\protect \APACyear {1995}}%
}]{%
zwaan1995event}
\APACinsertmetastar {%
zwaan1995event}%
\begin{APACrefauthors}%
Zwaan, R\BPBI A.%
, Langston, M\BPBI C.%
\BCBL {}\ \BBA {} Graesser, A\BPBI C.%
\end{APACrefauthors}%
\unskip\
\newblock
\APACrefYearMonthDay{1995}{}{}.
\newblock
{\BBOQ}\APACrefatitle {The Construction of Situation Models in Narrative Comprehension: An Event-Indexing Model} {The construction of situation models in narrative comprehension: An event-indexing model}.{\BBCQ}
\newblock
\APACjournalVolNumPages{Psychological Science}{6}{5}{292-297}.
\newblock
\begin{APACrefDOI} \doi{10.1111/j.1467-9280.1995.tb00513.x} \end{APACrefDOI}
\PrintBackRefs{\CurrentBib}

\end{thebibliography}

\end{document}